\documentclass[runningheads]{llncs}

\usepackage[utf8]{inputenc}

\usepackage{color,xcolor}
\usepackage{epsfig}
\usepackage{graphicx}
\usepackage{subfiles}

\usepackage{adjustbox}
\usepackage{array}
\newcolumntype{?}[1]{!{\vrule width #1}}
\usepackage{booktabs}
\usepackage{colortbl}
\usepackage{float,wrapfig}
\usepackage{hhline}
\usepackage{multirow}
\usepackage{subcaption} 
\usepackage[labelfont={bf},labelsep={period},font={small}]{caption}

\let\llncssubparagraph\subparagraph
\let\subparagraph\paragraph
\usepackage[compact]{titlesec}
\let\subparagraph\llncssubparagraph

\usepackage{amsmath,amsfonts,amssymb}
\usepackage{bm}
\usepackage{nicefrac}
\usepackage{microtype}

\usepackage{changepage}
\usepackage{extramarks}
\usepackage{fancyhdr}
\usepackage{lastpage}
\usepackage{setspace}
\usepackage{soul}
\usepackage{xspace}

\usepackage{url}

\usepackage{algorithm}
\usepackage{algpseudocode}
\usepackage{enumerate}

\newcolumntype{L}[1]{>{\raggedright\let\newline\\\arraybackslash\hspace{0pt}}m{#1}}
\newcolumntype{C}[1]{>{\centering\let\newline\\\arraybackslash\hspace{0pt}}m{#1}}
\newcolumntype{R}[1]{>{\raggedleft\let\newline\\\arraybackslash\hspace{0pt}}m{#1}}

\newcommand{\fig}[1]{Figure~\ref{#1}}
\newcommand{\tbl}[1]{Table~\ref{#1}}

\newcommand{\ignore}[1]{}

\renewcommand*{\thefootnote}{\fnsymbol{footnote}}

\makeatletter
\DeclareRobustCommand\onedot{\futurelet\@let@token\@onedot}
\def\@onedot{\ifx\@let@token.\else.\null\fi\xspace}

\def\eg{\emph{e.g}\onedot} 
\def\ie{\emph{i.e}\onedot} 
 
\def\etc{\emph{etc}\onedot} 
 
\def\etal{\emph{et al}\onedot}
\makeatother

\definecolor{MyDarkBlue}{rgb}{0,0.08,1}
\definecolor{MyDarkGreen}{rgb}{0.02,0.6,0.02}
\definecolor{MyDarkRed}{rgb}{0.8,0.02,0.02}
\definecolor{MyDarkOrange}{rgb}{0.40,0.2,0.02}
\definecolor{MyPurple}{RGB}{111,0,255}
\definecolor{MyRed}{rgb}{1.0,0.0,0.0}
\definecolor{MyGold}{rgb}{0.75,0.6,0.12}
\definecolor{MyDarkgray}{rgb}{0.66, 0.66, 0.66}

\algnewcommand{\LeftComment}[1]{\Statex \(\triangleright\) #1}

\newcommand{\stitle}[1]{\noindent\textbf{#1}}
\newcommand{\cifar}{CIFAR-10~\cite{krizhevsky2009learning}}

\newcommand{\Tensordecom}{Tensor factorization}

\newcommand{\blockqnn}{Block-QNN~\cite{zhong2017practical}}
\newcommand{\floplimit}{resource-constrained}
\newcommand{\acclimit}{accuracy-guaranteed}
\newcommand{\AccLimit}{Accuracy-Guaranteed}

\newcommand{\action}{\textit{a}}
\newcommand{\rewarda}{R_{err}}
\newcommand{\rewardb}{R_\text{FLOPs}}
\newcommand{\ours}{AMC (ours)}
\newcommand{\AMC}{AMC\xspace}
\newcommand{\AMCverbose}{AutoML for Model Compression}
\newcommand{\androidhalf}{1.95}
\newcommand{\adcfive}{285}

\begin{document}

\pagestyle{headings}
\mainmatter
\def\ECCV18SubNumber{1478}

\title{\AMC: \AMCverbose\ \\ and Acceleration on Mobile Devices}


\author{
Yihui He$^{\ddagger*}$ \and
Ji Lin$^{\dagger*}$ \and
Zhijian Liu$^\dagger$ \and
Hanrui Wang$^\dagger$ \and
Li-Jia Li$^\updownarrow$ \and
Song Han$^\dagger$
\email{\{jilin, songhan\}@mit.edu} \\
\ \\
$^\dagger$Massachusetts Institute of Technology \\
$^\ddagger$Carnegie Mellon University \\
$^\updownarrow$Google
}

\authorrunning{Yihui He, Ji Lin, Zhijian Liu, Hanrui Wang, Li-Jia Li and Song Han}
\titlerunning{\AMC: \AMCverbose\ \\ and Acceleration on Mobile Devices}

\renewcommand*{\thefootnote}{\fnsymbol{footnote}}
\footnotetext{$*$ Indicates equal contributions. Published as a conference paper in ECCV'18.} 


\maketitle

\begin{abstract} 
Model compression is a critical technique to efficiently deploy neural network models on mobile devices which have limited computation resources and tight power budgets. Conventional model compression techniques rely on hand-crafted heuristics and \emph{rule-based} policies that require domain experts to explore the large design space trading off among model size, speed, and accuracy, which is usually sub-optimal and time-consuming. In this paper, we propose \AMCverbose\ (\AMC) which leverage reinforcement learning to provide the model compression policy. This \emph{learning-based} compression policy outperforms conventional \emph{rule-based} compression policy by having higher compression ratio, better preserving the accuracy and freeing human labor. Under 4$\times$ FLOPs reduction, we achieved \textbf{2.7\%} better accuracy than the hand-crafted model compression policy for VGG-16 on ImageNet. We applied this automated, push-the-button compression pipeline to MobileNet and achieved \textbf{1.81$\times$} speedup of measured inference latency on an Android phone and \textbf{1.43$\times$} speedup on the Titan XP GPU, with only 0.1\% loss of ImageNet Top-1 accuracy.
\end{abstract}

\section{Introduction}
\begin{figure}[t]
\centering
\includegraphics[width=\linewidth]{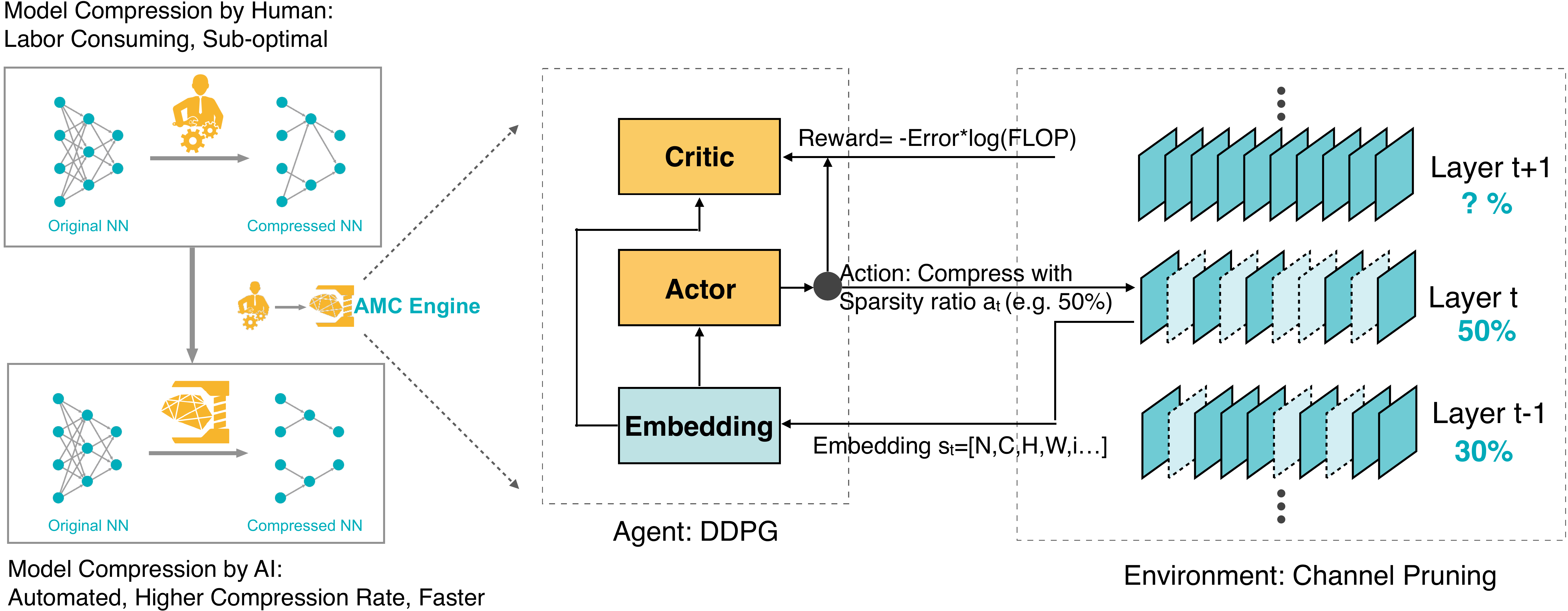}
\caption{Overview of \AMCverbose~(\AMC) engine. Left: \AMC~replaces human and makes model compression fully automated while performing better than human. Right: Form \AMC\ as a reinforcement learning problem. We process a pre-trained network (\eg, MobileNet) in a layer-by-layer manner. Our reinforcement learning agent (DDPG) receives the embedding $s_t$ from a layer $t$, and outputs a sparsity ratio $a_t$. After the layer is compressed with $a_t$, it moves to the next layer $L_{t+1}$. The accuracy of the pruned model with all layers compressed is evaluated. Finally, as a function of accuracy and FLOP, reward $R$ is returned to the reinforcement learning agent.}
\label{fig:net}
\end{figure}

In many machine learning applications such as robotics, self-driving cars, and advertisement ranking, deep neural networks are constrained by latency, energy and model size budget. Many approaches have been proposed to improve the hardware efficiency of neural networks by model compression~\cite{jaderberg2014speeding,han2015deep,he2017channel}. The core of model compression technique is to determine the compression policy for each layer as they have different redundancy, which conventionally requires hand-crafted heuristics and domain expertise to explore the large design space trading off among model size, speed, and accuracy. The design space is so large that human heuristic is usually sub-optimal, and manual model compression is time-consuming. To this end, we aim to automatically find the compression policy for an arbitrary network to achieve even better performance than human designed, rule-based model compression methods. 

Previously there were many rule-based model compression heuristics \cite{han2015learning,han_thesis}. For example, prune less parameters in the first layer which extracts low level features and have the least amount of parameters; prune more parameters in the FC layer since FC layers have the most parameters; prune less parameters in the layers that are sensitive to pruning, etc. 
However, as the layers in deep neural networks are not independent, these rule-based pruning policies are non-optimal, and doesn't transfer from one model to another model. Neural network architectures are evolving fast, and we need an automated way to compress them to improve engineer efficiency. As the neural network becomes deeper, the design space has exponential complexity , which is infeasible to be solved by greedy, rule-based methods. Therefore, 
we propose AutoML for Model Compression (\AMC) which leverages reinforcement learning to automatically sample the design space and  improve the model compression quality. \fig{fig:net} illustrates our \AMC engine. When compressing a network, our \AMC engine automates this process by learning-based policies, rather than relying on rule-based policy and experienced engineers.

We observe that the accuracy of the compressed model is very sensitive to the sparsity of each layer, requiring a fine-grained action space. Therefore, instead of searching over a discrete space, we come up with a continuous compression ratio control strategy with a DDPG~\cite{lillicrap2015continuous} agent to learn through trial and error: penalizing accuracy loss while encouraging model shrinking and speedup. The actor-critic structure also helps to reduce variance, facilitating stabler training. Specifically, our DDPG agent processes the network in a layer-wise manner. For each layer $L_t$, the agent receives a layer embedding $s_t$ which encodes useful characteristics of this layer, and then it outputs a precise compression ratio $a_t$. After layer $L_t$ is compressed with $a_t$, the agent moves to the next layer $L_{t+1}$. The validation accuracy of the pruned model with all layers compressed is evaluated without fine-tuning, which is an efficient delegate of the fine-tuned accuracy. This simple approximation can improve the search time not having to retrain the model, and provide high quality search result.
After the policy search is done, the best-explored model is fine-tuned to achieve the best performance.

We proposed two compression policy search protocols for different scenarios. For \emph{latency-critical} AI applications (\eg, mobile apps, self-driving cars, and advertisement rankings), we propose \emph{\floplimit} compression to achieve the best accuracy given the maximum amount of hardware resources (\eg, FLOPs, latency, and model size), 
For \emph{quality-critical} AI applications (\eg, Google Photos) where latency is not a hard constraint, we propose \emph{\acclimit} compression to achieve the smallest model with no loss of accuracy.  

We achieve \emph{\floplimit} compression by constraining the search space, in which the action space (pruning ratio) is constrained such that the model compressed by the agent is always below the resources budget. For \emph{\acclimit} compression, we define a reward that is a function of both accuracy and hardware resource. With this reward function, we are able to explore the limit of compression without harming the accuracy of models.

To demonstrate the wide and general applicability, we evaluate our \AMC engine on multiple neural networks, including VGG~\cite{simonyan2014very}, ResNet~\cite{he2016deep}, and MobileNet~\cite{howard2017mobilenets}, and we also test the generalization ability of the compressed model from classification to object detection. Extensive experiments suggest that \AMC offers better performance than hand-crafted heuristic policies. For ResNet-50, we push the expert-tuned compression ratio~\cite{han_thesis} from 3.4$\times$ to \textbf{5$\times$} with no loss of accuracy. Furthermore, we reduce the FLOPs of MobileNet~\cite{howard2017mobilenets} by 2$\times$, achieving top one accuracy of 70.2\%, which is on a better Pareto curve than 0.75 MobileNet, and we achieve a speedup of \textbf{1.53$\times$} on the Titan XP and \textbf{\androidhalf$\times$} on an Android phone.

\begin{table}[t]
\centering
\caption{Comparisons of reinforcement learning approaches for models searching (NAS: Neural Architecture Search~\cite{zoph2017learning}, NT: Network Transformation~\cite{cai2017reinforcement}, N2N: Network to Network~\cite{ashok2017n2n}, and \AMC: \AMCverbose. \AMC distinguishes from other works by getting reward without fine-tuning, continuous search space control, and can produce both \acclimit \ and hardware \floplimit\ models.}
\label{tab:compare}

\begin{tabular}{lC{1cm}C{1cm}C{1cm}C{1cm}}
\toprule
& NAS & NT & N2N & \textbf{\AMC} \\
\midrule
optimize for accuracy & \checkmark & \checkmark & \checkmark & \checkmark \\
optimize for latency & & & & \checkmark \\
simple, non-RNN controller &  & & & \checkmark\\
fast exploration with few GPUs & & \checkmark & \checkmark & \checkmark \\
continuous action space & & & & \checkmark \\

\bottomrule
\end{tabular}

\end{table}

\section{Related Work}
\stitle{CNN Compression and Acceleration.} 
Extensive works~\cite{han2015learning,han2015deep,luo2017thinet,dong2017more,han2016eie,han2017ese} have been done on accelerating neural networks by compression. Quantization~\cite{zhu2016trained,courbariaux2016binarynet,rastegari2016xnor} and special convolution implementations~\cite{mathieu2013fast,vasilache2014fast,lavin2015fast,bagherinezhad2016lcnn} can also speed up the neural networks.
\Tensordecom~\cite{lebedev2014speeding,gong2014compressing,kim2015compression,Masana_2017_ICCV} decomposes weights into light-weight pieces, for example~\cite{xue2013restructuring,denton2014exploiting,girshick2015fast} proposed to accelerate the fully connected layers with truncated SVD; Jaderberg~\etal~\cite{jaderberg2014speeding} proposed to factorize layers into 1$\times$3 and 3$\times$1; and Zhang~\etal~\cite{zhang2016accelerating} proposed to factorize layers into 3$\times$3 and 1$\times$1. Channel pruning~\cite{polyak2015channel,hu2016network,anwar2016compact,molchanov2016pruning} removes the redundant channels from feature maps. A common problem of these methods is how to determine the sparsity ratio for each layer.

\stitle{Neural Architecture Search and AutoML.}
Many works on searching models with reinforcement learning and genetic algorithms~\cite{stanley2002evolving,real2017large,brock2017smash,miikkulainen2017evolving} greatly improve the performance of neural networks. NAS~\cite{zoph2017learning} aims to search the transferable network blocks, and its performance surpasses many \textit{manually} designed architectures~\cite{szegedy2015going,he2016deep,chollet2016xception}. Cai~\etal~\cite{cai2017reinforcement} proposed to speed up the exploration via network transformation~\cite{chen2015net2net}. Inspired by them, N2N~\cite{ashok2017n2n} integrated reinforcement learning into channel selection. In \tbl{tab:compare}, we demonstrate several merits of our \AMC engine. Compared with previous work, AMC engine optimizes for both accuracy and latency, requires a simple non-RNN controller, can do fast exploration with fewer GPU hours, and also support continuous action space.

\section{Methodology}

We present an overview of our \AMCverbose (\AMC) engine in \fig{fig:net}. We aim to automatically find the redundancy for each layer, characterized by sparsity. We train an reinforcement learning agent to predict the action and give the sparsity, then perform form the pruning. We quickly evaluate the accuracy after pruning but before fine-tuning as an effective delegate of final accuracy. Then we update the agent by encouraging smaller, faster and more accurate models.

\subsection{Problem Definition}

Model compression is achieved by reducing the number of parameters and computation of each layer in deep neural networks. There are two categories of pruning: fine-grained pruning and structured pruning. 
\emph{Fine-grained pruning}~\cite{han2015deep} aims to prune individual unimportant elements in weight tensors, which is able to achieve very high compression rate with no loss of accuracy. However, such algorithms result in an irregular pattern of sparsity, and it requires specialized hardware such as EIE \cite{han2016eie}
for speed up. 
\emph{Coarse-grained / structured pruning}~\cite{li2016pruning} aims to prune entire regular regions of weight tensors (\eg, channel, row, column, block, \etc). The pruned weights are regular and can be accelerated directly with off-the-shelf hardware and libraries. Here we study structured pruning that shrink the input channel of each convolutional and fully connected layer.

Our goal is to precisely find out the effective \emph{sparsity} for each layer, which used to be manually determined in previous studies~\cite{molchanov2016pruning,li2016pruning,he2017channel}. Take convolutional layer as an example. The shape of a weight tensor is $n\times c\times  k \times k$, where $n, c$ are output and input channels, and $k$ is the kernel size. For fine-grained pruning, the sparsity is defined as the number of zero elements divided by the number of total elements, \ie $\text{\#zeros} / (n \times c \times k \times h)$. For channel pruning, we shrink the weight tensor to $n\times c^\prime\times  k \times k$ (where $c^\prime < c$), hence the sparsity becomes $c^\prime / c$.

\subsection{Automated Compression with Reinforcement Learning}

AMC leverages reinforcement learning for efficient search over action space. Here we introduce the detailed setting of reinforcement learning framework.

\subsubsection{The State Space}\label{sec:state}
For each layer $t$, we have 11 features that characterize the state $s_t$: 
\begin{align}
(t, n, c, h, w, stride, k, FLOPs[t], reduced, rest, \action_{t-1})
\end{align}
where $t$ is the layer index, the dimension of the kernel is $n\times c\times k\times k$, and the input is $c\times h \times w$. $FLOPs[t]$ is the FLOPs of layer $L_t$. $Reduced$ is the total number of reduced FLOPs in previous layers. $Rest$ is the number of remaining FLOPs in the following layers. Before being passed to the agent, they are scaled within $[0, 1]$. Such features are essential for the agent to distinguish one convolutional layer from another.

\subsubsection{The Action Space}\label{sec:action}
Most of the existing works use discrete space as coarse-grained action space (\eg, $\{64, 128, 256, 512\}$ for the number of channels). 
Coarse-grained action space might not be a problem for a high-accuracy model architecture search. However, we observed that model compression is very sensitive to sparsity ratio and requires fine-grained action space, leading to an explosion of the number of discrete actions (Sec.~\ref{sec:exp:vgg}). Such large action spaces are difficult to explore efficiently~\cite{lillicrap2015continuous}. Discretization also throws away the order: for example, 10\% sparsity is more aggressive than 20\% and far more aggressive than 30\%.

As a result, we propose to use continuous action space $\action \in (0, 1]$, which enables more fine-grained and accurate compression.

\subsubsection{DDPG Agent}\label{sec:agent}
As illustrated in \fig{fig:net}, the agent receives an embedding state $s_t$ of layer $L_t$ from the environment and then outputs a sparsity ratio as action $\action_t$. The underlying layer is compressed with $\action_t$ (rounded to the nearest feasible fraction) using a specified compression algorithm (\eg, channel pruning). Then the agent moves to the next layer $L_{t+1}$, and receives state $s_{t+1}$.
After finishing the final layer $L_T$, the reward accuracy is evaluated on the validation set and returned to the agent. For fast exploration, we evaluate the reward accuracy without fine-tuning, which is a good approximation for fine-tuned accuracy (Sec.~\ref{sec:acc}).

We use the deep deterministic policy gradient (DDPG) for continuous control of the compression ratio, which is an off-policy actor-critic algorithm. For the exploration noise process, we use truncated normal distribution:

\begin{align}
\mu^\prime(s_t) \sim \text{TN}\left(\mu\left(s_t \mid \theta_t^\mu\right), \sigma^2, 0, 1\right)
\end{align}

During exploitation, noise $\sigma$ is initialized as $0.5$ and is decayed after each episode exponentially.

Following \blockqnn, which applies a variant form of Bellman's Equation~\cite{watkins1989learning}, each transition in an episode is $(s_t, a_t, R, s_{t+1})$, where $R$ is the reward after the network is compressed. During the update, the baseline reward $b$ is subtracted to reduce the variance of gradient estimation, which is an exponential moving average of the previous rewards~\cite{zoph2016neural,cai2017reinforcement}:
\begin{equation}
\begin{gathered}
Loss = \frac{1}{N} \sum_i\left(y_i - Q\left(s_i, a_i \mid \theta^Q\right)\right)^2 \\
y_i = r_i - b + \gamma Q(s_{i+1}, \mu(s_{i+1})\mid \theta^Q)
\end{gathered}
\end{equation}
The discount factor $\gamma$ is set to 1 to avoid over-prioritizing short-term rewards~\cite{baker2016designing}.

\subsection{Search Protocols}

\subsubsection{Resource-Constrained Compression}\label{sec:wall}

By limiting the action space (the sparsity ratio for each layer), we can accurately arrive at the target compression ratio. Following~\cite{zoph2017learning,baker2016designing,zhong2017practical}, we use the following reward:
\begin{align}
\rewarda = -Error
\end{align}
This reward offers no incentive for model size reduction, so we the achieve target compression ratio by an alternative way: limiting the action space. 
Take fine-grained pruning for model size reduction as an example:
we allow arbitrary action $\action$ at the first few layers; we start to limit the action $\action$ when we find that the budget is insufficient even after compressing \textbf{all} the following layers with most aggressive strategy. Algorithm~\ref{Alg1} illustrates the process. (For channel pruning, the code will be longer but similar, since removing input channels of layer $L_t$ will also remove the corresponding output channels of $L_{t-1}$, reducing parameters/FLOPs of both layers). Note again that our algorithm is not limited to constraining \textit{model size} and it can be replaced by other resources, such as FLOPs or the actual inference time on mobile device. Based on our experiments (Sec.~\ref{sec:cifarresource}),
as the agent receives no incentive for going below the budget, it can precisely arrive at the target compression ratio.

\begin{algorithm}[t]

\caption{Predict the sparsity ratio $\texttt{action}_t$ for layer $L_t$ with constrained model size (number of parameters) using fine-grained pruning}

\label{Alg1}
\begin{algorithmic}

\LeftComment{Initialize the reduced model size so far}
\If {$t$ is equal to $0$}
	\State $\texttt{W}_\text{reduced} \gets 0 $
\EndIf
\State
\LeftComment{Compute the agent's action and bound it with the maximum sparsity ratio}
\State $\texttt{action}_t \gets \mu^{\prime}(s_t)$
\State $\texttt{action}_t \gets \min(\texttt{action}_t, \texttt{action}_\text{max})$
\State
\LeftComment{Compute the model size of the whole model and all the later layers}
\State $\texttt{W}_\text{all} \gets \sum_k \texttt{W}_k$
\State $\texttt{W}_\text{rest} \gets \sum_{k=t+1} \texttt{W}_k$
\State
\LeftComment{Compute the number of parameters we have to reduce in the current layer if all the later layers are pruned with the maximum sparsity ratio. $\alpha$ is the target sparsity ratio of the whole model.}
\State $\texttt{W}_\text{duty} \gets \alpha \cdot \texttt{W}_\text{all} - \texttt{action}_\text{max} \cdot \texttt{W}_\text{rest} - \texttt{W}_\text{reduced}$
\State
\LeftComment{Bound $\texttt{action}_t$ if it is too small to meet the target model size reduction}
\State $\texttt{action}_t \gets \max(\texttt{action}_t, \texttt{W}_\text{duty} / \texttt{W}_t)$
\State
\LeftComment{Update the accumulation of reduced model size}
\State $\texttt{W}_\text{reduced} \gets \texttt{W}_\text{reduced} + \texttt{action}_t \cdot \texttt{W}_t$
\State
\State \textbf{return} $\texttt{action}_t$
\end{algorithmic}
\end{algorithm}

\subsubsection{\AccLimit\ Compression}\label{sec:reward}
By tweaking the reward function, we can accurately find out the limit of compression that offers no loss of accuracy. 
We empirically observe that $Error$ is inversely-proportional to $\log(FLOPs)$ or $\log(\#Param)$~\cite{canziani2016analysis}. Driven by this, we devise the following reward function:
\begin{align}
\rewardb = - Error \cdot \log(\text{FLOPs}) \\
R_\text{Param} = -Error \cdot \log(\text{\#Param})
\end{align}
This reward function is sensitive to $Error$; in the meantime, it provides a small incentive for reducing $\text{FLOPs}$ or model size.
Based on our experiments in \fig{sec:cifarperform}, we note that our agent automatically finds the limit of compression.

\section{Experimental Results}
For fine-grained pruning~\cite{han2015deep}, we prune the weights with least magnitude. The maximum sparsity ratio $a_{max}$ for convolutional layers is set to 0.8, and $a_{max}$ for fully connected layer is set to 0.98. For channel pruning, we use \textit{max response} selection (pruning the weights according to the magnitude~\cite{han2015learning}), and preserve Batch Normalization~\cite{ioffe2015batch} layers during pruning instead of merging them into convolutional layers. The maximum sparsity ratios $a_{max}$ for all layers are set to 0.8. Note that the manual upper bound $a_{max}$ is only intended for faster search, one can simply use $a_{max}=1$ which also produces similar results. Our actor network $\mu$ has two hidden layers, each with 300 units. The final output layer is a sigmoid layer to bound the actions within $(0, 1)$. Our critic network $Q$ also had two hidden layers, both with 300 units. Actions arr included in the second hidden layer. We use $\tau = 0.01$ for the soft target updates and train the network with 64 as batch size and 2000 as replay buffer size. Our agent first explores 100 episodes with a constant noise $\sigma=0.5$, and then exploits 300 episodes with exponentially decayed noise $\sigma$.

\subsection{CIFAR-10 and Analysis}
We conduct extensive experiments and fully analyze our \AMC\ on \cifar to verify the effectiveness of the 2 search protocols. CIFAR dataset consists of 50k training and 10k testing $32\times32$ tiny images in ten classes. We split the training images into 45k/5k train/validation. 
The accuracy reward is obtained on validation images. Our approach is computationally efficient: the RL can finish searching within \textbf{1 hour} on a single GeForce GTX TITAN Xp GPU.

\begin{table}[t]
\setlength{\tabcolsep}{3pt}
\centering\small
\caption{Pruning policy comparison of Plain-20, ResNets~\cite{he2016deep} on \cifar. $R_\text{Err}$ corresponds to FLOPs-constrained compression with channel pruning, while $R_\text{Param}$  corresponds to accuracy guaranteed compression with fine-grained pruning. For both shallow network Plain-20 and deeper network ResNets, \AMC outperforms \textit{hand-crafted} policies by a large margin.  This enables efficient exploration without fine-tuning. Although \AMC makes many trials on model architecture, we have separate validation and test dataset. No over-fitting is observed.}

\begin{tabular}{cccccc}
\toprule
Model & Policy & Ratio & Val Acc. & Test Acc. & Acc. after FT. \\
\midrule
\multirow{4}{*}{\begin{tabular}[c]{@{}c@{}}
Plain-20\\(90.5\%)
\end{tabular}}
& deep (handcraft) & \multirow{4}{*}{\begin{tabular}[c]{@{}c@{}}50\% FLOPs \end{tabular}}  & 79.6 & 79.2 & 88.3 \\
& shallow (handcraft) &  & 83.2 & 82.9 & 89.2 \\
& uniform (handcraft) &   & 84.0 & 83.9 & 89.7 \\           
& \textbf{AMC ($R_\text{Err})$}  &  & \textbf{86.4} & \textbf{86.0} & \textbf{90.2} \\

\midrule
\multirow{3}{*}{\begin{tabular}[c]{@{}c@{}}ResNet-56\\ (92.8\%)\end{tabular}}
& uniform (handcraft) & \multirow{3}{*}{\begin{tabular}[c]{@{}c@{}}50\% FLOPs \end{tabular}}  & 87.5 & 87.4 & 89.8 \\
& deep (handcraft) &   & 88.4 & 88.4 & 91.5 \\
& \textbf{AMC ($R_\text{Err})$} &   & \textbf{90.2} & \textbf{90.1} & \textbf{91.9} \\ 
\midrule
\begin{tabular}{c} ResNet-50 \\ (93.53\%) \end{tabular} 
& \textbf{AMC ($R_\text{Param})$} & 60\% Params & \textbf{93.64} & \textbf{93.55} & -\\

\bottomrule
\end{tabular}

\label{tab:cifarpr}

\end{table}

\begin{figure}[t]
\centering
\includegraphics[width=\linewidth]{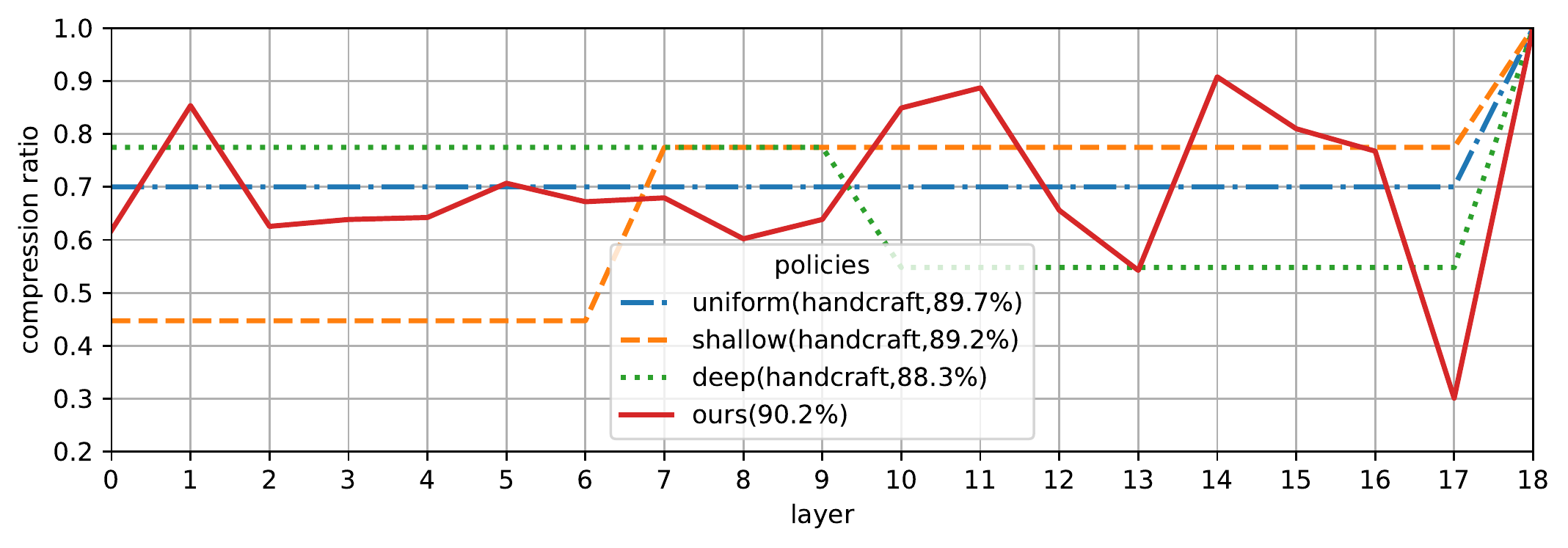}
\caption{Comparisons of pruning strategies for Plain-20 under $2\times$. \textit{Uniform} policy sets the same compression ratio for each layer uniformly. \textit{Shallow} and \textit{deep} policies aggressively prune shallow and deep layers respectively. Policy given by \AMC\ looks like sawtooth, which resembles the bottleneck architecture~\cite{he2016deep}. The accuracy given by AMC outperforms hand-crafted policies. (\textit{better viewed in color})}
\label{fig:cifarpr}
\end{figure}

\subsubsection{FLOPs-Constrained Compression.}\label{sec:cifarresource}

We conducted FLOPs-constrained experiments on CIFAR-10 with channel pruning. We compare our approach with three empirical policies~\cite{li2016pruning,he2017channel} illustrated in \fig{fig:cifarpr}: \textit{uniform} sets compression ratio uniformly, \textit{shallow} and \textit{deep} aggressively prune shallow and deep layers respectively. Based on sparsity distribution of different networks, a different strategy might be chosen. 

In \tbl{tab:cifarpr}, we show us using reward $R_{err}$ to accurately find the sparsity ratios for pruning 50\% for Plain-20 and ResNet-56~\cite{he2016deep} and compare it with empirical policies. We outperform empirical policies by a large margin. The best pruning setting found by AMC differs from hand-crafted heuristic (\fig{fig:cifarpr}). It learns a bottleneck architecture~\cite{he2016deep}.

\subsubsection{\AccLimit\ Compression.}\label{sec:cifarperform}
By using the $R_\text{Param}$ reward, our agent can automatically find the limit of compression, with smallest model size and little loss of performance. As shown in  Table~\ref{tab:cifarpr}, we compress ResNet-50 with fine-grained pruning on CIFAR-10. 
The result we obtain has up to 60\% compression ratio with even a little higher accuracy on both validation set and test set, which might be in light of the regularization effect of pruning. 

Since our reward $R_{\text{Param}}$ focuses on $Error$ and offers very little incentive to compression in the meantime, it prefers the high-performance model with harmless compression. To shorten the search time, we obtain the reward using validation accuracy without fine-tuning. We believe if reward were fine-tuned accuracy, the agent should compress more aggressively, because the fine-tuned accuracy is much closer to the original accuracy.

\subsubsection{Speedup policy exploration.}\label{sec:acc}

Fine-tuning a pruned model usually takes a very long time. We observe a correlation between the pre-fine-tune accuracy and the post fine-tuning accuracy~\cite{han2015learning,he2017channel}. As shown in Table~\ref{tab:cifarpr}, policies that obtain higher validation accuracy correspondingly have higher fine-tuned accuracy. This enables us to predict final model accuracy without fine-tuning, which results in an efficient and faster policy exploration.

The validation set and test set are separated, and we only use the validation set to generate reward during reinforcement learning. In addition, the compressed models have fewer parameters. As shown in Table~\ref{tab:cifarpr}, the test accuracy and validation accuracy are very close, indicating no over-fitting.

\subsection{ImageNet}\label{exp:imagenet}

On ImageNet, we use 3000 images from the \textbf{training set} to evaluate the reward function in order to prevent over-fitting. 
The latency is measured with $224\times224$ input throughout the experiments.

\subsubsection{Push the Limit of Fine-grained Pruning.}

\begin{figure}[t]
\centering
\includegraphics[height=4.6cm]{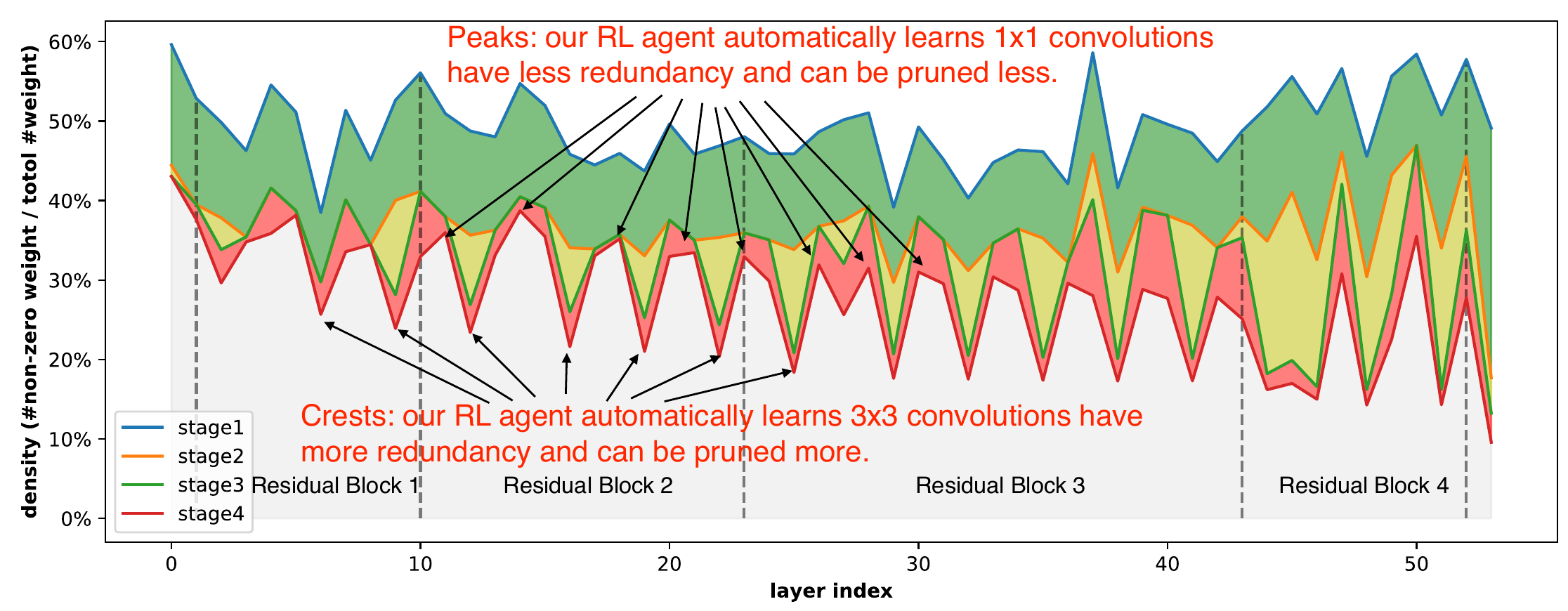}
\caption{ The pruning policy (sparsity ratio) given by our reinforcement learning agent for ResNet-50. With 4 stages of iterative pruning, we find very salient sparsity pattern across layers: peaks are $1\times1$ convolution, crests are $3\times3$ convolution. \textbf{The reinforcement learning agent automatically learns that $3\times3$ convolution has more redundancy than $1\times1$ convolution and can be pruned more.} }
\label{fig:stage4}
\end{figure} 

\begin{figure}[t]
\centering
\includegraphics[width=0.93\linewidth]{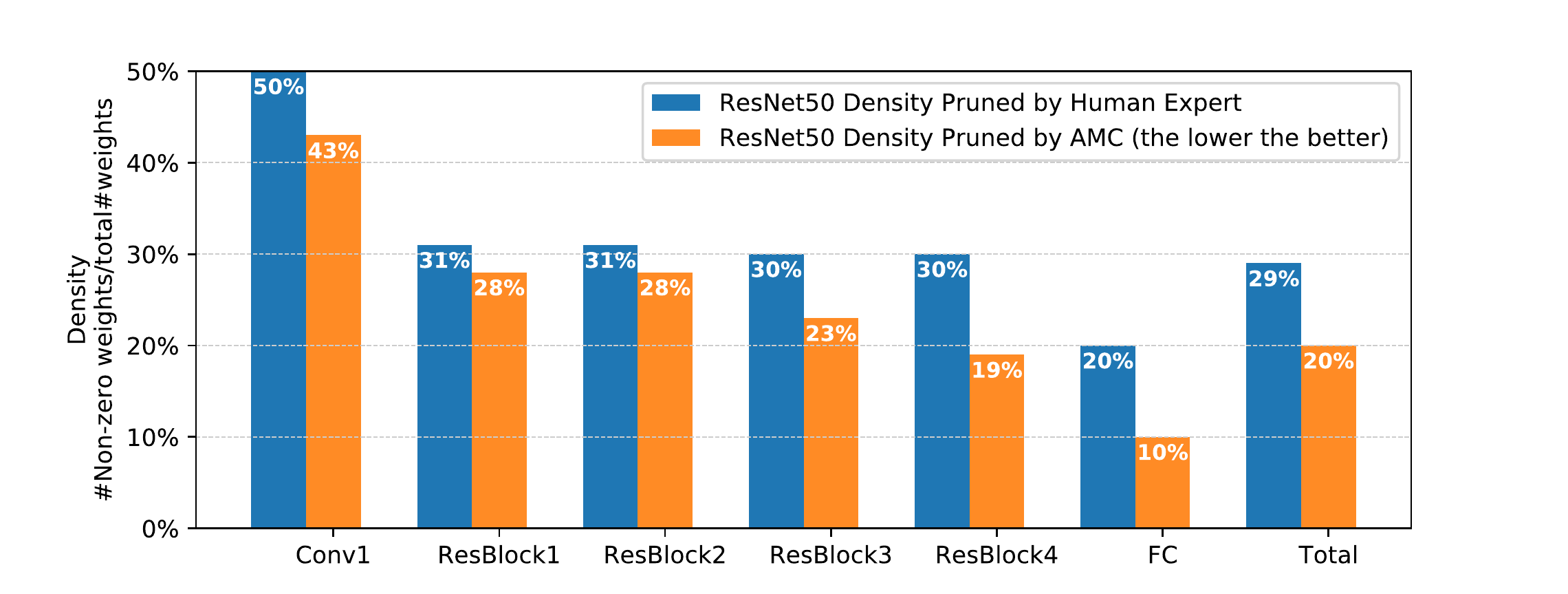}
\caption{Our reinforcement learning agent (\AMC) can prune the model to a lower density compared with human experts without losing accuracy. (Human expert: $3.4\times$ compression on ResNet50. \AMC\ : $5\times$ compression on ResNet50.)}
\label{fig:finegrain}
\end{figure}

Fine-grained pruning method prunes neural networks based on individual connections to achieve sparsity in both weights and activations, which is able to achieve higher compression ratio and can be accelerated with specialized hardware such as~\cite{han2016eie,han2017ese,parashar2017scnn}. However, it requires iterative prune \& fine-tune procedure to achieve decent performance~\cite{han2015learning}, and single-shot pruning without retraining will greatly hurt the prediction accuracy with large compression rate (say $4\times$), which cannot provide useful supervision for reinforcement learning agent.

To tackle the problem, we follow the settings in \cite{han_thesis} to conduct 4-iteration pruning \& fine-tuning experiments, where the overall density of the full model is set to [50\%, 35\%, 25\% and 20\%] in each iteration. For each stage, we run \AMC\ to determine the sparsity ratio of each layer given the overall sparsity. The model is then pruned and fine-tuned for 30 epochs following common protocol. With that framework, we are able to push the expert-tuned compression ratio of ResNet-50 on ImageNet from $3.4\times$ to $\mathbf{5\times}$ (see Figure~\ref{fig:finegrain}) without loss of performance on ImageNet (original ResNet50's [top-1, top-5] accuracy=[76.13\%, 92.86\%]; \AMC\ pruned model's accuracy=[76.11\%, 92.89\%]). 
The density of each layer during each stage is displayed in Figure~\ref{fig:stage4}. The peaks and crests show that the RL agent automatically learns to prune $3\times3$ convolutional layers with larger sparsity, since they generally have larger redundancy; while prunes more compact $1\times1$ convolutions with lower sparsity. 
The density statistics of each block is provided in Figure~\ref{fig:finegrain}. We can find that the density distribution of \AMC is quite different from human expert's result shown in Table 3.8 of \cite{han_thesis}, suggesting that AMC can fully explore the design space and allocate sparsity in a better way.

\subsubsection{Comparison with Heuristic Channel Reduction.}\label{sec:exp:vgg}

\begin{table}[t]
\centering
\caption{Learning-based model compression (AMC) outperforms rule-based model compression. Rule-based heuristics are suboptimal. (for reference, the baseline top-1 accuracy of VGG-16 is 70.5\%, MobileNet is 70.6\%, MobileNet-V2 is 71.8\%).}
\label{tab:vggde}

\begin{tabular}{c|c|c|c}
\hline
 & policy & FLOPs & $\Delta Acc~\%$ \\ \hline
\multirow{5}{*}{\begin{tabular}[c]{@{}c@{}}VGG-16 \end{tabular}}  
& FP (handcraft)~\cite{li2016pruning} & \multirow{5}{*}{20\%} & -14.6    \\ 
\cline{2-2} \cline{4-4}
& RNP (handcraft)~\cite{lin2017runtime} &  & -3.58    \\ \cline{2-2} \cline{4-4}
& SPP (handcraft)~\cite{wang2017structured} &  & -2.3    \\ \cline{2-2} \cline{4-4} 
& CP (handcraft)~\cite{he2017channel} &   & -1.7 \\ \cline{2-2} \cline{4-4} 
& \textbf{\ours} &      & -\textbf{1.4}     \\ \hline \hline
\multirow{4}{*}{MobileNet} 
& uniform (0.75-224)~\cite{howard2017mobilenets}  &  56\%  & -2.5     \\ 
\cline{2-4} 
& \textbf{\ours}& 50\% & \textbf{-0.4}  \\
\cline{2-4} 
& uniform (0.75-192)~\cite{howard2017mobilenets} & 41\% & -3.7 \\
\cline{2-4} 
& \textbf{\ours}& 40\% & \textbf{-1.7}     \\

\hline \hline
\multirow{2}{*}{\ MobileNet-V2\ } 
& uniform (0.75-224)~\cite{sandler2018inverted}  &  
\multirow{2}{*}{\begin{tabular}[c]{@{}c@{}}70\% \end{tabular}}  
& -2.0     \\ 
\cline{2-2}  \cline{4-4} 
& \textbf{\ours}& & \textbf{-1.0}  \\
\hline
\end{tabular}
\end{table}

Here we compare  AMC with existing state-of-the-art channel reduction methods: FP~\cite{li2016pruning}, RNP~\cite{lin2017runtime} and SPP~\cite{wang2017structured}. All the methods proposed a heuristic strategy to design the pruning ratio of each layer. FP~\cite{li2016pruning} proposed a sensitive analysis scheme to estimate the sensitivity of each layer by evaluating the accuracy with single layer pruned. Layers with lower sensitivity are pruned more aggressively. Such method assumes that errors of different pruned layers can be summed up linearly, which does not stand according to our experiments. RNP~\cite{lin2017runtime} groups all convolutional channels into 4 sets and trains a RL agent to decide on the 4 sets according to input image. However, the action space is very rough (only 4 actions for each layer), and it cannot reduce the model size. SPP~\cite{wang2017structured} applies PCA analysis to each layer and takes the reconstruction error as the sensitivity measure to determine the pruning ratios. Such analysis is conducted based on one single layer, failing to take the correlations between layers into consideration. 
We also compare our method to the original channel pruning paper (CP in Table~\ref{tab:vggde}), in which the sparsity ratios of pruned VGG-16~\cite{simonyan2014very} are carefully tuned by human experts (\verb|conv5| are skipped, sparsity ratio for \verb|conv4| and remaining layers is $1:1.5$). The results of pruned VGG-16 are presented in Table~\ref{tab:vggde}. Consistent with our CIFAR-10 experiments (Sec.~\ref{sec:cifarresource}),  AMC outperforms all heuristic methods by more than \textbf{0.9\%}, and beats human expert by \textbf{0.3\%} without any human labor.

Apart from VGG-16, we also test  AMC on modern efficient neural networks MobileNet~\cite{howard2017mobilenets} and MobileNet-V2~\cite{sandler2018inverted}. Since the networks have already been very compact, it is much harder to further compress them. The easiest way to reduce the channels of a model is to use uniform channel shrinkage, \ie use a width multiplier to uniformly reduce the channels of each layer with a fixed ratio. Both MobileNet and MobileNet-V2 present the performance of different multiplier and input sizes, and we compare our pruned result with models of same computations. The format are denoted as \emph{uniform (depth multiplier - input size)}. We can find that our method consistently outperforms the uniform baselines. Even for the current state-of-the-art efficient model design MobileNet-V2,   AMC can still improve its accuracy by $1.0\%$ at the same computation~(Table \ref{tab:vggde}). The pareto curve of MobileNet is presented in Figure~\ref{fig:sub1}.

\subsubsection{Speedup Mobile Inference.}

\begin{figure}[t]
\centering
\begin{subfigure}{.5\textwidth}
  \centering
  \includegraphics[width=1\linewidth]{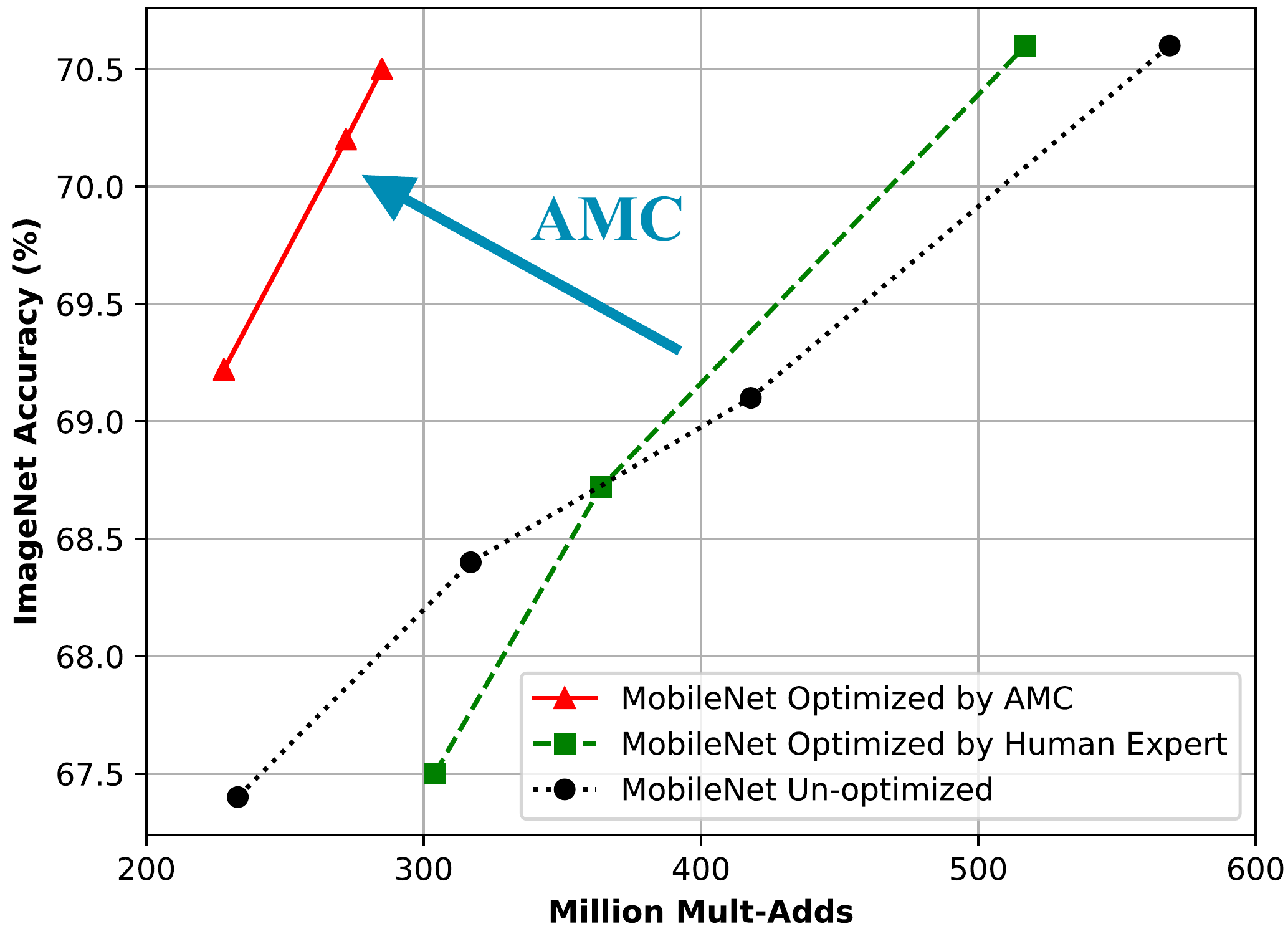}
  \caption{Accuracy v.s. MACs}
  \label{fig:sub1}
\end{subfigure}%
\begin{subfigure}{.49\textwidth}
  \centering
  \includegraphics[width=1\linewidth]{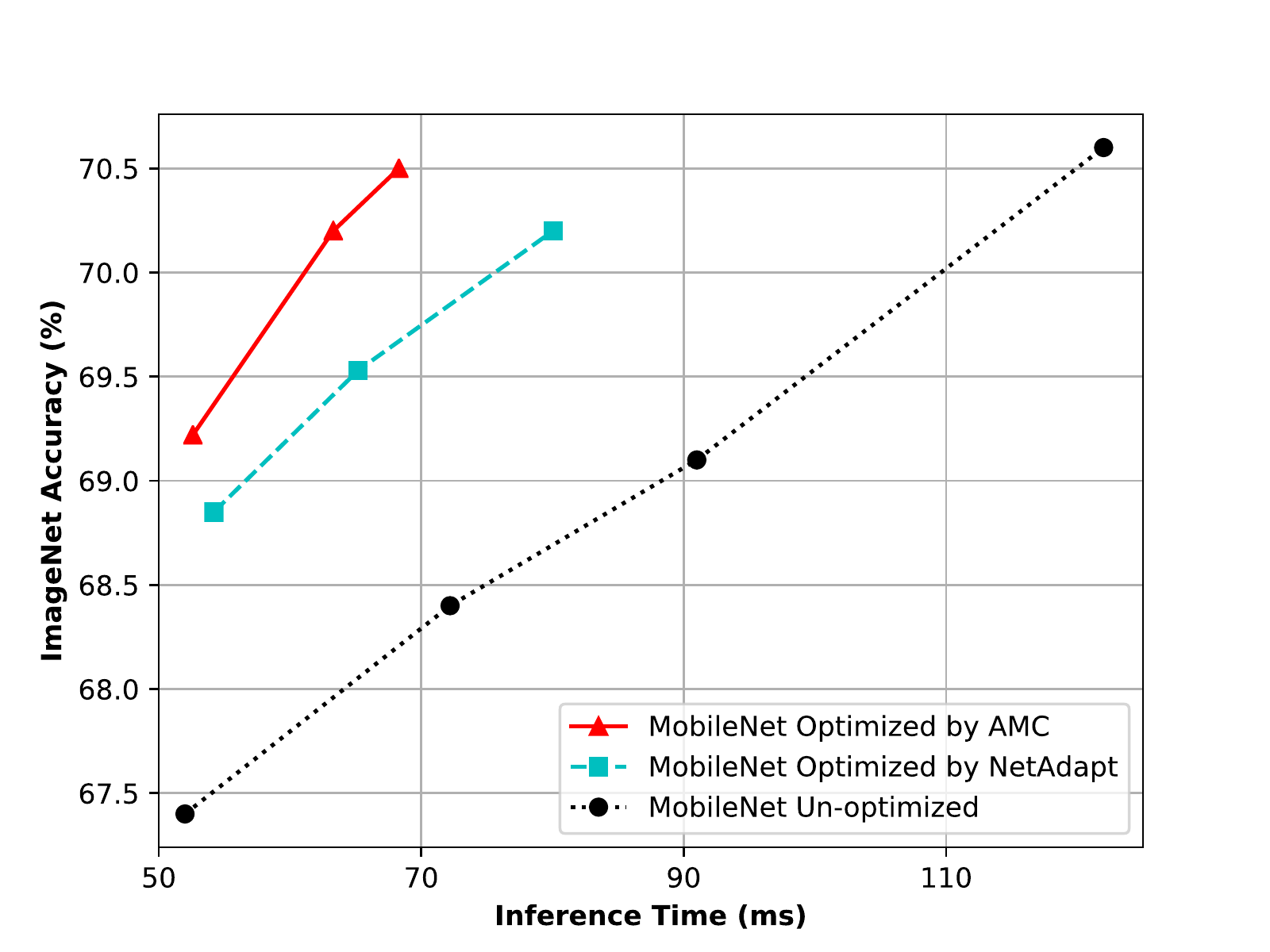}
  \caption{Accuracy v.s. Inference time}
  \label{fig:sub2}
\end{subfigure}
\caption{ (a) Comparing the accuracy and MAC trade-off among \AMC, human expert, and unpruned MobileNet. \AMC strictly dominates human expert in the pareto optimal curve.  (b) Comparing the accuracy and latency trade-off among \AMC, NetAdapt, and unpruned MobileNet. \AMC significantly improves the pareto curve of MobileNet. Reinforcement-learning based AMC surpasses heuristic-based NetAdapt on the pareto curve (inference time both measured on Google Pixel 1).}
\label{fig:test}
\end{figure}

\begin{table}[t]
\setlength{\tabcolsep}{3pt}
\centering 
\caption{AMC speeds up MobileNet. Previous attempts using \textit{rule-based} policy to prune MobileNet lead to significant accuracy degradation~\cite{li2016pruning} while \AMC use \textit{learning-based} pruning which well preserve the accuracy. 
On Google Pixel-1 CPU, \AMC achieves  $1.95\times$ measured speedup with batch size one, while saving the memory by 34\%. On NVIDIA Titan XP GPU, \AMC achieves $1.53\times$ speedup with batch size of 50. The input image size is 224$\times$224 for all experiments and no quantization is applied for apple-to-apple comparison.}

\scalebox{0.91}{
\begin{tabular}{c|c|cc|cc|ccc}
\hline
\multirow{2}{*}{} & \multirow{2}{*}{\begin{tabular}[c]{@{}c@{}}Million\\ MAC\end{tabular}} & \multirow{2}{*}{\begin{tabular}[c]{@{}c@{}}top-1\\ acc.\end{tabular}} & \multirow{2}{*}{\begin{tabular}[c]{@{}c@{}}top-5\\ acc.\end{tabular}} & \multicolumn{2}{c|}{GPU} & \multicolumn{3}{c}{Android} \\
 &  &  &  & \begin{tabular}[c]{@{}c@{}}latency \\\end{tabular} & \begin{tabular}[c]{@{}c@{}}speed \\\end{tabular} & 
\begin{tabular}[c]{@{}c@{}}latency \\\end{tabular} & \begin{tabular}[c]{@{}c@{}}speed \\\end{tabular} & \begin{tabular}[c]{@{}c@{}}memory \\\end{tabular} \\ \hline\hline

\begin{tabular}[c]{@{}c@{}}100\%\\ MobileNet\end{tabular} & 569 & 70.6\% & 89.5\% & 0.46ms & 2191 fps &  123.3ms & 8.1 fps & 20.1MB \\ 
\begin{tabular}[c]{@{}c@{}}75\%\\ MobileNet\end{tabular} & 325 & 68.4\% & 88.2\% & 0.34ms & 2944 fps &  72.3ms &13.8 fps & 14.8MB \\ \hline 
NetAdapt \cite{yang2018netadapt} & - & 69.8\% & - & - & - & 70.0ms & 14.3 fps & - \\\hline
\begin{tabular}[c]{@{}c@{}}\textbf{AMC} \\ ($50\%$ FLOPs)\\ \end{tabular} & 285 & 70.5\% & 89.3\% & 0.32ms & \begin{tabular}[c]{@{}c@{}}\textbf{3127 fps}\\ (\textbf{1.43}$\times$)\end{tabular} &  68.3ms & \begin{tabular}[c]{@{}c@{}}\textbf{14.6 fps}\\ (\textbf{1.81}$\times$)\end{tabular} & 14.3MB \\ 
\begin{tabular}[c]{@{}c@{}}\textbf{AMC} \\ ($50\%$ Latency) \end{tabular} & 272 & 70.2\% & 89.2\% & 0.30ms & \begin{tabular}[c]{@{}c@{}}\textbf{3350 fps}\\ (\textbf{1.53}$\times$)\end{tabular} & 63.3ms & \begin{tabular}[c]{@{}c@{}}\textbf{16.0 fps}\\ (\textbf{1.95}$\times$)\end{tabular}  & 13.2MB \\
\hline

\end{tabular}  
}

\label{tab:android}
\end{table}

Mobile inference acceleration has drawn people's attention in recent years. 
Not only can AMC optimize FLOPs and model size, it can also optimize the inference latency, directly benefiting mobile developers. For all mobile inference experiments, we use TensorFlow Lite framework for timing evaluation.

We prune MobileNet~\cite{howard2017mobilenets}, a highly compact network consisting of depth-wise convolution and point-wise convolution layers, and measure how much we can improve its inference speed. Previous attempts using \textit{hand-crafted} policy to prune MobileNet led to significant accuracy degradation~\cite{li2016pruning}: pruning MobileNet to 75.5\% original parameters results in 67.2\% top-1 accuracy\footnote{http://machinethink.net/blog/compressing-deep-neural-nets/}, which is even worse than the original 0.75 MobileNet (61.9\% parameters with 68.4\% top-1 accuracy). However, our \AMC pruning policy significantly improves pruning quality: on ImageNet, \AMC-pruned MobileNet achieved 70.5\% Top-1 accuracy with \adcfive\ MFLOPs, compared to the original 0.75 MobileNet's 68.4\% Top-1 accuracy with 325 MFLOPs.
As illustrated in Figure~\ref{fig:sub1}, human expert's hand-crafted policy achieves slightly worse performance than that of the original MobileNet under $2\times$ FLOPs reduction. However, with \AMC, we significantly raise the pareto curve, improving the accuracy-MACs trade-off of original MobileNet.

By substituting FLOPs with latency, we can change from FLOPs-constrained search to latency-constrained search and directly optimize the inference time. Our experiment platform is Google Pixel 1 with Qualcomm Snapdragon 821 SoC. As shown in Figure~\ref{fig:sub2}, we greatly reduce the inference time of MobileNet under the same accuracy. We also compare our learning based policy with a heuristic-based policy~\cite{yang2018netadapt}, and AMC better trades off accuracy and latency. Furthermore, since AMC uses the validation accuracy before fine-tuning as the reward signal while \cite{yang2018netadapt} needs local fine-tuning after each step,  AMC is more sampling-efficient, requiring fewer GPU hours for policy search. 

We show the detailed statistics of our pruned model in Table~\ref{tab:android}. Model searched with 0.5$\times$ FLOPs and 0.5$\times$ inference time are profiled and displayed. 
For 0.5$\times$ FLOPs setting, we achieve \textbf{1.81}$\times$ speed up on a Google Pixel 1 phone, and for 0.5$\times$ FLOPs setting, we accurately achieve \textbf{1.95}$\times$ speed up, which is very close to actual 2$\times$ target, showing that AMC can directly optimize inference time and achieve accurate speed up ratio.
We achieve 2.01$\times$ speed up for 1$\times$1 convolution but less significant speed up for depth-wise convolution due to the small computation to communication ratio. \AMC compressed models also consumes less memory.
On GPUs, we also achieve up to 1.5$\times$ speedup, less than mobile phone, which is because a GPU has higher degree of parallelism than a mobile phone.

\subsubsection{Generalization Ability.}
\begin{table}[t]
\centering
\caption{Compressing Faster R-CNN with VGG16 on PASCAL VOC 2007. Consistent with classification task, \AMC also results in better performance under the same compression ratio on object detection task.}
\label{tab:detection}

\begin{tabular}{c|c|c}
\hline
                                                   & mAP (\%)             & mAP [0.5, 0.95] (\%)   \\ \hline
baseline                                           & 68.7                 & 36.7                 \\ \hline
$2\times$ handcrafted~\cite{he2017channel}         & 68.3 (-0.4)          & 36.7 (-0.0)          \\ \hline
$4\times$ handcrafted~\cite{he2017channel}         & 66.9 (-1.8)          & 35.1 (-1.6)          \\ \hline
$4\times$ handcrafted~\cite{zhang2016accelerating} & 67.8 (-0.9)          & 36.5 (-0.2)          \\ \hline
$4\times$ \ours                                    & \textbf{68.8 (+0.1)} & \textbf{37.2 (+0.5)} \\ \hline
\end{tabular}
\end{table}

We evaluate the generalization ability of \AMC on PASCAL VOC object detection task~\cite{pascal-voc-2007}. We use the compressed VGG-16 (from Sec \ref{sec:exp:vgg}) as the backbone for Faster R-CNN~\cite{ren2015faster}. 
In \tbl{tab:detection}, \AMC\ achieves \textbf{0.7\%} better mAP[0.5, 0.95]) than the best hand-crafted pruning methods under the same compression ratio. \AMC\ even surpasses the baseline by \textbf{0.5\%} mAP. 
We hypothesize this improvement as that the optimal compression policy found by the RL agent also serves as effective regularization.

\section{Conclusion}
Conventional model compression techniques use hand-crafted features and require domain experts to explore a large design space and trade off between model size, speed, and accuracy, which is usually suboptimal and labor-consuming. In this paper, we propose \AMCverbose\ (\AMC), which leverages reinforcement learning to automatically search the design space, greatly improving the model compression quality. We also design two novel reward schemes to perform both \floplimit\ compression and \acclimit\ compression.
Compelling results have been demonstrated for MobileNet, MobileNet-V2, ResNet and VGG on Cifar and ImageNet. The compressed model generalizes well from classification to detection tasks. On the Google Pixel 1 mobile phone, we push the inference speed of MobileNet from 8.1 fps to 16.0 fps. \AMC facilitates efficient deep neural networks design on mobile devices.

\section*{Acknowledgements}
We thank Quoc Le, Yu Wang and Bill Dally for the supportive feedback. 
We thank Jiacong Chen for drawing the cartoon on the left of Figure 1.  

\clearpage

\bibliographystyle{splncs04}
\bibliography{egbib}


\end{document}


\newpage
\begin{center}
\textbf{\large Supplementary Materials}
\end{center}

\section{Comparison with Random Search}
\begin{figure}
\centering
\includegraphics[height=6.5cm]{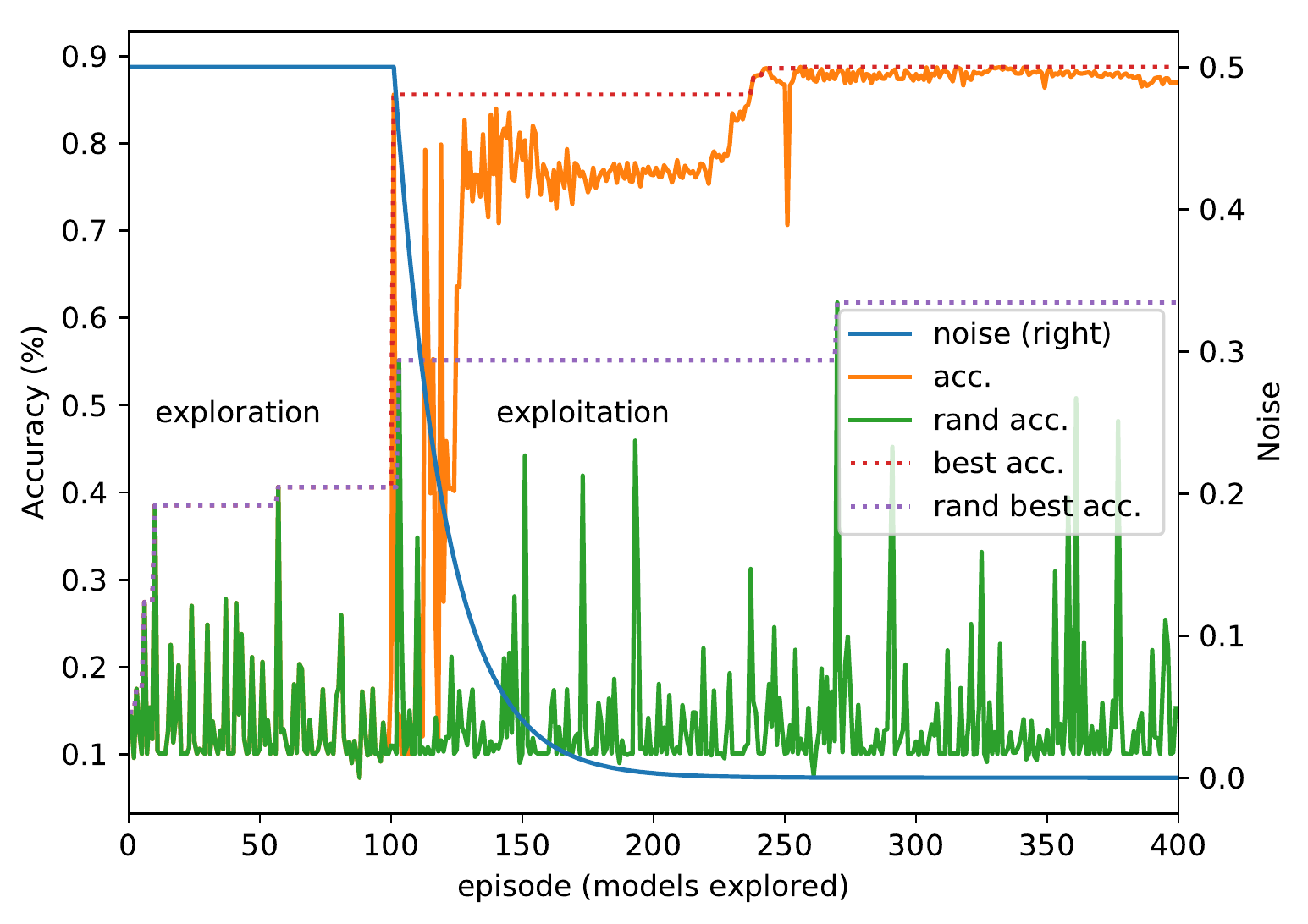}
\caption{Validation accuracy and noise decay versus episode, using spatial decomposition on CIFAR-10 under $2\times$. We explore for 100 episodes with constant noise $\sigma$. Then we exploit for 300 episodes with exponentially decayed noise $\sigma$. Notice that best accuracy from random exploration plateaus for very long time. Random search cannot find good enough results with limited episodes (\textit{better viewed in color)}
  }
\label{fig:loss}
\end{figure}

Theoretically, after a large number of trials, the random search result will be close to the reinforcement learning result. It is unclear how many trials are necessary for the random search to succeed, especially for small networks on CIFAR-10. As shown in \fig{fig:loss}, we search sparsity policy for spatial decomposition of Plain-20 under $2\times$. Before 100 episodes, our \AMC behaves the same as random search. Both explore with constant noise $\sigma=0.5$. After 100 episodes, our \AMC exploits with exponentially decayed noise $\sigma$ and start learning. It can find a good model quickly; however, random search plateaus. 400 episodes is insufficient for the random search to get good enough results, even for a small network on CIFAR-10.

\section{Effectiveness of Layer Embeddings}\label{sec:exp:state}
\begin{figure}
\centering
\includegraphics[height=6.5cm]{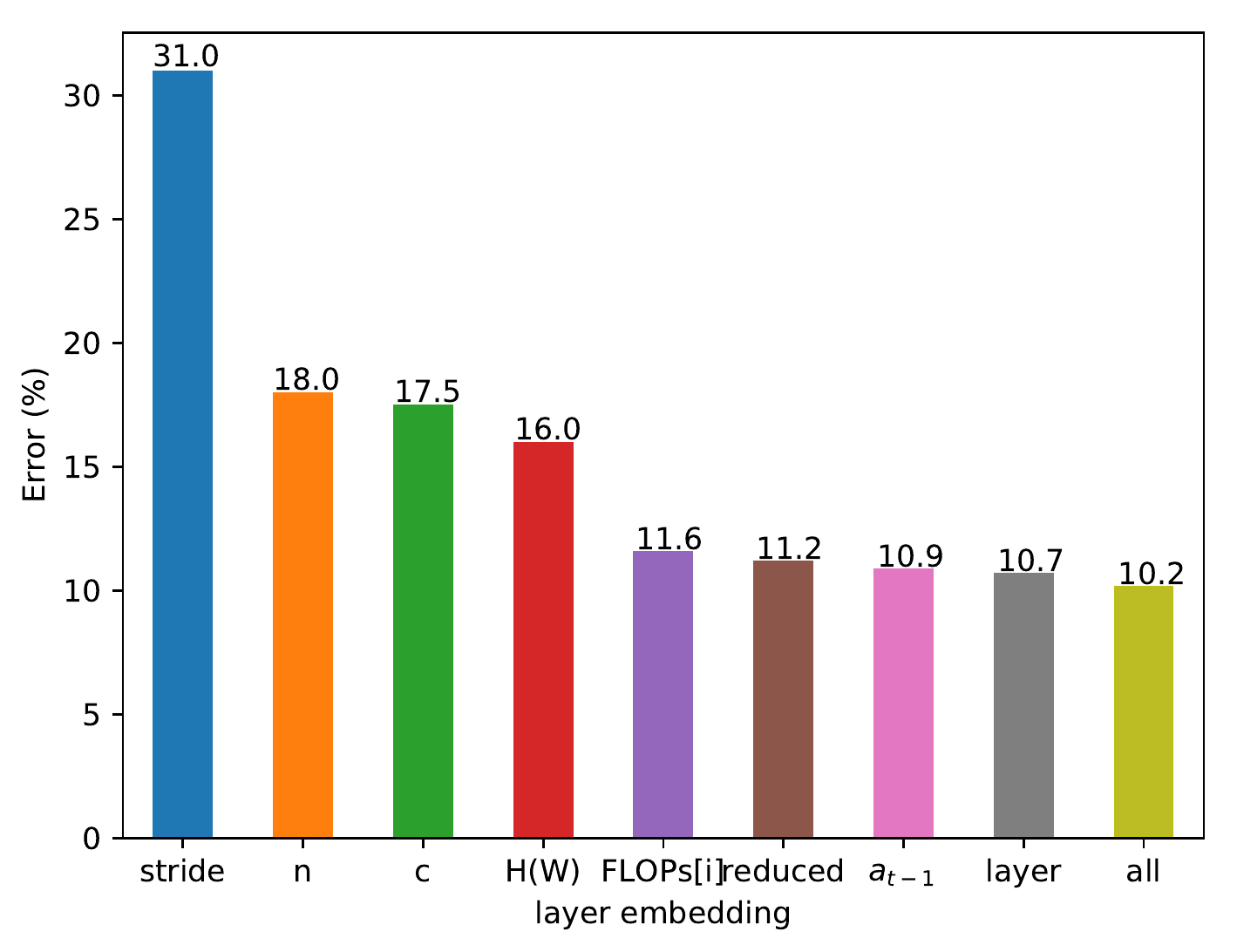}
\caption{Effectiveness of each layer embedding, evaluated with spatial decomposition for Plain-20 $2\times$. Combining all embeddings together, the performance is even better (\textit{smaller is better})}
\label{fig:state}
\end{figure}

Layer embedding state $s_t$ (Sec.~\ref{sec:state}) is essential for \AMC to distinguish different convolutional layers. As shown in \fig{fig:state}, we evaluate the effectiveness of each layer's embedding, and the combination of all of them with spatial decomposition Plain-20 under $2\times$. 
If we just use $stride$ as the layer embedding, it is not informative enough to distinguish different layers. The performances of $n, c, H, W$ are similar, since their values change together when $stride$ is 2. $FLOPs[i], reduced, a_{t-1}$ and layer index are most effective, since they dynamically change at each layer. $a_{t-1}$ and $reduced$ resemble the RNN used by other architecture search approaches~\cite{zoph2017learning,cai2017reinforcement}, since $s_t$ is affected by $s_{t-1}$. In the following experiments, we combine all these embeddings, which makes the performance even better.

\section{More speed-up results: Samsung Galaxy S7}

We provide the speed-up obtained by AMC measured on a Samsung Galaxy S7 phone with Qualcomm Snapdragon 820 SoC in Table~\ref{tab:samsung}.

\begin{table}

\setlength{\tabcolsep}{3pt}
\centering 
\caption{AMC speeds up MobileNet on Samsung Galaxy S7 with TF-Lite.}
\label{tab:samsung}
\scalebox{0.9}{
\begin{tabular}{c|c|cc|ccccc}
\hline
\multirow{2}{*}{} & \multirow{2}{*}{\begin{tabular}[c]{@{}c@{}}Million\\ MAC\end{tabular}} & \multirow{2}{*}{\begin{tabular}[c]{@{}c@{}}top-1\\ acc.\\ (\%)\end{tabular}} & \multirow{2}{*}{\begin{tabular}[c]{@{}c@{}}top-5\\ acc.\\ (\%)\end{tabular}} & \multicolumn{5}{c}{Android} \\
 &  &  &   & \begin{tabular}[c]{@{}c@{}}conv.\\ (ms)\end{tabular} & \begin{tabular}[c]{@{}c@{}}depth.\\ (ms)\end{tabular} & \begin{tabular}[c]{@{}c@{}}total\\ (ms)\end{tabular} & \begin{tabular}[c]{@{}c@{}}speed\\ \\\end{tabular} & \begin{tabular}[c]{@{}c@{}}memory\\ \\\end{tabular} \\ \hline\hline

\begin{tabular}[c]{@{}c@{}}1.0\\ MobileNet\end{tabular} & 569 & 70.9 & 89.5  & 102.3 & 12.0 & 119.0 & \begin{tabular}[c]{@{}c@{}}8.4 fps\\ (1$\times$)\end{tabular} & 20.1MB \\
\begin{tabular}[c]{@{}c@{}}0.75\\ MobileNet\end{tabular} & 325 & 68.4 & 88.2 & 56.0 & 9.9 & 69.5 & \begin{tabular}[c]{@{}c@{}}14.4 fps\\ (1.71$\times$)\end{tabular} & 14.8MB \\ \hline 
\begin{tabular}[c]{@{}c@{}}$0.5\times$ FLOPs\\ \textbf{\ours}\end{tabular} & 285 & 70.5 & 89.3  & 51.7 & 9.2 & 64.4 & \begin{tabular}[c]{@{}c@{}}\textbf{15.5 fps}\\ (\textbf{1.85}$\times$)\end{tabular} & 14.3MB \\ 
\begin{tabular}[c]{@{}c@{}}$0.5\times$ Time\\ \textbf{\ours}\end{tabular} & 272 & 70.2 & 89.2  & 48.7 & 8.3 & 59.7 & \begin{tabular}[c]{@{}c@{}}\textbf{16.8 fps}\\ (\textbf{2.00}$\times$)\end{tabular}  & 13.2MB \\
\hline
\end{tabular}
}
\end{table}